\documentclass[11pt,a4paper]{article}
\usepackage[hyperref]{acl2017}

\usepackage{pbox}
\usepackage{times}
\usepackage{latexsym}
\usepackage{color}
\usepackage{times}
\usepackage{bbm}
\usepackage{dsfont}
\usepackage{qtree}
\usepackage{multirow}
\usepackage{url}
\usepackage{latexsym}
\usepackage{graphicx}
\usepackage{amsmath,amssymb,amsthm}

\newcommand{\R}{\ensuremath{\mathbb{R}}}

\newcommand{\paren}[1]{\left(#1\right)}

\usepackage{url}
\usepackage{diagbox}
\usepackage{fancyvrb}

\aclfinalcopy 

\title{Toward Continual Learning for Conversational Agents}

\author{Sungjin Lee \\
  Microsoft Research / Redmond, WA, USA \\
  {\tt sule@microsoft.com} 
}

\date{}

\begin{document}
\maketitle
\begin{abstract}
While end-to-end neural conversation models have led to promising advances in reducing hand-crafted features and errors induced by the traditional complex system architecture, they typically require an enormous amount of data due to the lack of modularity. Previous studies adopted a hybrid approach with knowledge-based components either to abstract out domain-specific information or to augment data to cover more diverse patterns. On the contrary, we propose to directly address the problem using recent developments in the space of continual learning for neural models. Specifically, we adopt a domain-independent neural conversational model and introduce a novel neural continual learning algorithm that allows a conversational agent to accumulate skills across different tasks in a data-efficient way. To the best of our knowledge, this is the first work that applies continual learning to conversation systems. We verified the efficacy of our method through a conversational skill transfer from either synthetic dialogs or human-human dialogs to human-computer conversations in a customer support domain. 
\end{abstract}

\section{Introduction}
\label{sec:intro}
Conversational bots become increasingly popular in a wide range of business areas. In order to support the rapid development of bots, a number of bot building platforms have been launched, for example, from Microsoft, Amazon and so on. 
Despite this progress, the development of a business-critical bot still requires a serious effort from the design to actual implementation of several components such as language understanding, state tracking, action selection, and language generation.  
Not only does this complexity prevent casual developers from building quality bots but also introduces an unavoidable degradation in performance due to some non-trivial problems including unclear state representation design, insufficient labeled data and error propagation down the pipeline.   

Recently, end-to-end (E2E) approaches using Deep Neural Networks (DNNs) have shown the potential to solve such problems -- the DNNs induce a latent representation in the course of the joint optimization of all components without requiring any labeling on internal state. Despite such appealing aspects, the neural E2E approaches also have major challenges to overcome. The state-of-the-art systems require numerous dialogs only to learn simple behaviors. In general, it is expensive to collect a sufficient amount of dialogs from a target task. A possibility to the data-intensiveness problem would be to repurpose already built models through fine-tuning with a few additional dialogs from the target task. For example, one can add a payment handling capability to a new bot by repurposing any model that is already trained on payment-related conversations. It has been shown, however, that neural models tend to forget what it previously learned when it continuously trains on a new task, which is what is called {\em Catastrophic Forgetting}~\cite{french1999catastrophic}. Another possible approach would be to compose only relevant parts from each of the pretrained models for the target task. Unfortunately, it is also unclear how to compose neural models due to the lack of modularity.  

There have been prior studies that partly address the data-intensiveness problem outside the neural model~\cite{wen2016network,williams2017hybrid,zhao2017generative,eshghi2017bootstrapping,wenmultinlg16,zhao2017generative}.
In this work, we instead propose to directly address the problem with recent developments in the space of continual learning for neural models. Specifically, we adopt a domain-independent neural conversational model and introduce a novel {\em Adaptive Elastic Weight Consolidation} (AEWC) algorithm to continuously learn a new task without forgetting valuable skills that are already learned. 
To test our method, we first continuously train a model on synthetic data that only consists of general conversation patterns like opening/closing turns and then on a corpus of human-computer (H-C) dialogs in a customer support domain that barely has general conversations. Then, we show that the resulting model is able to handle both general and task-specific conversations without forgetting general conversational skills. Second,
we continuously train a model on a large amount of human-human (H-H) dialogs and then on a small number of H-C dialogs in the same customer support domain. As H-H conversations typically cover various out-of-domain topics including general conversations, this allows us to show that the resulting model can carry out the target task while handling general conversations that do not occur in the H-C dialogs.

The rest of this paper is organized as follows. In Section~\ref{sec:related} we present a brief summary of related work. In Section~\ref{sec:method} we describe our approach for continual learning for conversational agents. In Section~\ref{sec:experiments} we discuss our experiments. We finish with conclusions and future work in Section~\ref{sec:conclusion}.

\section{Related Work}
\label{sec:related}
\paragraph{Conversational Systems}
Traditionally, a dialog system has a pipeline architecture, typically consisting of language understanding, dialog state tracking, dialog control policy, and language generation~\cite{jokinen2009spoken,young2013pomdp}. With this architecture, developing dialog systems requires designing the input and output representation of multiple components. It also, oftentimes, involves writing a large number of handcrafted rules and laboriously labeling dialog datasets to train statistical components. In order to avoid this costly manual endeavor, a line of research has emerged to introduce end-to-end trainable neural models~\cite{sordoni2015neural,vinyals2015neural,serban2016building,wen2016network,bordes2016learning}. But the E2E neural approach is data-intensive requiring over thousands of dialogs to learn even simple behaviors. 
Broadly there are two lines of work addressing the data-intensiveness problem. The first makes use of domain-specific information and linguistic knowledge to abstract out the data~\cite{wen2016network,williams2017hybrid,zhao2017generative,eshghi2017bootstrapping}.
The second line of work adopts data recombination approaches to generate counterfeit data that mimics target domain dialogs~\cite{wenmultinlg16,zhao2017generative}.
The prior approaches, however, partly bring us back to the downside of traditional approaches: The difficulty in maintenance increases as the number of rules grows; The system quality depends on external expertise; It is hard to scale out over different domains. Furthermore, the increased size of training data, as a result of data recombination approaches, would lead to a significant increase in training time. Unlike prior work, we approach the problem from the standpoint of continual learning where a single neural network accumulates task-relevant knowledge over time and exploits this knowledge to rapidly learn a new task from a small number of examples.

\paragraph{Continual Learning for Neural Networks}
Previous studies that address the catastrophic forgetting problem are broadly partitioned into three groups. First, architectural approaches reduce interference between tasks by altering the architecture of the network. The simplest form is to copy the entire network for the previous task and add new features for a new task.~\cite{rusu2016progressive,lee2016dual}. Though this prevents forgetting on earlier tasks, the architectural complexity grows with the number of tasks. ~\newcite{fernando2017pathnet} proposes a single network where a subset of different modules gets picked for each task based on an evolutionary idea to alleviate the complexity issue.
Second, functional approaches encourage similar predictions for the previous and new tasks. \newcite{li2016learning} applies the old network to the training data of the new task and uses the output as pseudo-labels. \newcite{jung2016less} performs a similar regularization on the distance between the final hidden activations. But the additional computation using the old network makes functional approaches computationally expensive.
Lastly, implicitly distributed knowledge approaches use neural networks of a large capacity to distribute knowledge for each task using dropout, maxout, or local winner-take-all~\cite{goodfellow2013empirical,srivastava2013compete}. But these earlier approaches had limited
success and failed to preserve performance on the old task when an extreme change to the environment occurred.  Recently, \newcite{kirkpatrick2017overcoming} proposed elastic weight consolidation (EWC) which makes use of a point estimate for the Fisher information metric as a weighting factor for a distance penalty between the parameters of the new and old tasks. To alleviate the cost of exactly computing the diagonal of the Fisher metric, \newcite{zenke2017continual} presented an online method that accumulates the importance of individual weights over the entire parameter trajectory during training. This method, however, could yield an inaccurate importance measure when the loss function is not convex. Thus, we propose an adaptive version of the online method that applies exponential decay to cumulative quantities.

\section{Continual Learning for Conversational Agents}
\label{sec:method}
In this section, we describe a task-independent conversation model and an adaptive online algorithm for continual learning which together allow us to sequentially train a conversation model over multiple tasks without forgetting earlier tasks.

\subsection{Task-Independent Conversation Model}
\label{conv_model}
As we need to use the same model structure across different tasks, including open-domain dialogs as well as task-oriented dialogs, we adopt a task-independent model.
Thus, our model should be able to induce a meaningful representation from raw observations without access to hand-crafted task-specific features.~\footnote{Inspired by~\citeauthor{williams2017hybrid}~(\citeyear{williams2017hybrid}), one can still make use of action masks without hurting the task independence.}
In order to achieve this goal, we employ a neural encoder for state tracking and a neural ranking model for action selection.

\paragraph{State Tracking}
\label{state_tracking}
Due to the sequential nature of the conversation, variants of Recurrent Neural Networks (RNNs)~\cite{medsker1999recurrent} have been widely adopted for modeling conversation. 
State tracking models with RNN architecture, however, usually tend to become less effective as the length of a sequence increases due to the gradual information loss along the recurring RNN units.  
To address this problem, we use a hierarchical recurrent encoder architecture which has been recently adopted for generative conversation models~\cite{serban2016building}.
Specifically, our model mimics the natural structure in language --- a conversation consists of a sequence of utterances, an utterance is a sequence of words, a word, in turn, is composed of characters.

In order to capture character-level patterns in word embeddings, we concatenate the word embeddings with the output of the bidirectional character-level RNN encoder. 
We use Long Short-Term Memory (LSTM)~\cite{hochreiter1997long} as the RNN unit
that takes an input vector $x$ and a state vector $h$ to output a new state vector $h' = \phi(x, h)$.
Let $\mathcal{C}$ denote the set of characters and $\mathcal{W}$ the set of words.
$w \in \mathcal{W}$ is a sequence of characters $(c_1 \ldots c_m) \in \mathcal{C}^m$. We compute the embedding of a word, $v$, as follows:
\begin{align*}
f^{\mathcal{C}}_i &= \phi^{\mathcal{C}}_f\paren{e_{c_i}, f^{\mathcal{C}}_{i-1}} &&\forall i = 1 \ldots m \\
b^{\mathcal{C}}_i &= \phi^{\mathcal{C}}_b\paren{e_{c_i}, b^{\mathcal{C}}_{i+1}} &&\forall i = m \ldots 1\\
v &= f^{\mathcal{C}}_m \oplus b^{\mathcal{C}}_1 \oplus e_w &&
\end{align*}
where $\oplus$ denote the vector concatenation operation. $e_c \in \R^{d_c}$ and $e_w \in \R^{d_w}$ are character embeddings and word embeddings, respectively.

\noindent
Next, we have another bidirectional LSTM-RNN layer that takes the word embeddings $(v_1 \ldots v_n)$ of an utterance $(w_1 \ldots w_n) \in \mathcal{W}^n$ to generate an utterance embedding $u \in \R^{d_u}$:
\begin{align} 
f^{\mathcal{W}}_i &= \phi^{\mathcal{W}}_f\paren{v_i, f^{\mathcal{W}}_{i-1}}  &&\forall i = 1 \ldots n \\
b^{\mathcal{W}}_i &= \phi^{\mathcal{W}}_b\paren{v_i, b^{\mathcal{W}}_{i+1}} &&\forall i = n \ldots 1 \\
u &= f^{\mathcal{W}}_n \oplus b^{\mathcal{W}}_1 \label{utt_emb}
\end{align}

\noindent
After that, we have a unidirectional LSTM-RNN layer that takes a sequence of pairs of user utterance embedding, $(u_1 \ldots u_t)$, and previous system action embeddings~\footnote{We treat system actions the same as user utterances and encode both with a common LSTM-RNN encoder}, $(a_1 \ldots a_t)$,  as input to induce state embeddings $(s_1 \ldots s_t)$:
\begin{align} \label{state_emb}
s_i &= \phi_s\paren{u_i \oplus a_i, s_{i-1}}  &&\forall i = 1 \ldots t
\end{align}

\paragraph{Action Ranking}
In a continual learning setting, there is no predefined action set from which a classifier selects an action across different tasks. Thus, we cast the action selection problem as a ranking problem where we consider the affinity strength between a state embedding $s$ (Eq.~\ref{state_emb}) and a set of candidate system action embeddings $\{a_i\}$ (Eq.~\ref{utt_emb}):~\footnote{We employ the same utterance-level encoder as in the state tracking to yield the embeddings of the candidate system actions.} 
\begin{equation}
\rho(a|s) = s^T M a
\end{equation}
where $M \in \R^{d_s} \times \R^{d_u}$ projects the state embedding onto the action space.
In order to optimize the projection matrix, $M$, we adopt the {\em Plackett-Luce} model~\cite{plackett1975analysis}. The Plackett-Luce model normalizes ranking score by transforming real-valued scores into a probability distribution:
\begin{equation}
p(a|s) = \frac{\exp(s^T M a)}{\sum_i \exp(s^T M a_i)}
\end{equation}
With the transformed probability distribution, we minimize the cross-entropy loss against the true label distribution. Thanks to the normalization, our ranking model performs a more effective penalization to negative candidates.

\subsection{Adaptive Elastic Weight Consolidation}
In order to achieve continual learning, we need to minimize the total loss function summed over all tasks, $\mathcal{L} = \Sigma_\mu L_\mu$, without access to the true loss functions of prior tasks. 
A catastrophic forgetting arises when minimizing $L_\mu$ leads to an undesirable increase of the loss on prior tasks $L_\nu$ with $\nu < \mu$.
Variants of the EWC algorithm tackle this problem by optimizing a modified loss function:
\begin{equation}
\label{surrogate_loss}
\tilde{L}_\mu = L_\mu + \underbrace{c\sum_k\Omega_k^\mu(\bar{\theta}_k-\theta_k)^2}_{\text{surrogate loss}}
\end{equation}
where $c$ represents an weighting factor between prior and current tasks, $\theta$ all model parameters introduced in Section~\ref{conv_model}, $\bar{\theta}_k$ the parameters at the end of the previous task and $\Omega_k^\mu$ regularization strength per parameter $k$. The bigger $\Omega_k^\mu$, the more influential is the parameter.
EWC defines $\Omega^\mu$, for example, to be a point estimate which is equal to the diagonal entries of the {\em Fisher information matrix} at the final parameter values. 
Since EWC relies on a point estimate, we empirically noticed that sometimes $\Omega^\mu$ fails to capture the parameter importance when the loss surface is relatively flat around the final parameter values as $\Omega^\mu$ essentially decreases to zero. 

In contrast to EWC, \citeauthor{zenke2017continual}~(\citeyear{zenke2017continual}) computes an importance measure online by taking the path integral of the change in loss along the entire trajectory through parameter space.
Specifically, the per-parameter contribution $\omega_k^\mu$ to changes in the total loss is defined as follows:
\begin{equation}
\omega_k^\mu = -\int_{t^{\mu-1}}^{t^\mu}g_k(\theta(t))\theta^\prime_k(t)dt
\end{equation}
where $\theta(t)$ is the parameter trajectory as a function of time $t$, $g(\theta) = \frac{\partial L}{\partial \theta}$ and $\theta^\prime_k(t) = \frac{\partial \theta}{\partial t}$. Note that the minus sign indicates that we are interested in decreasing the loss. In practice, we can approximate $\omega_k^\mu$ as the sum of the product of the gradient $g_k(t)$ with the parameter update $\Delta_k(t)$. Having defined $\omega_k^\mu$, $\Omega_k^\mu$ is defined such that the regularization term carries the same units as the loss by dividing $\omega_k^\mu$ by the total displacement in parameter space:
\begin{equation}
\Omega_k^\mu = \sum_{\nu < \mu}\frac{\omega_k^\nu}{(\Delta_k^\nu)^2 + \zeta}
\end{equation}
where $\Delta_k^\nu$ quantifies how far the parameter moved during the training process for task $\nu$. $\zeta$ is introduced to keep the expression from exploding in cases where the denominator gets close to zero. Note that, with this definition, the quadratic surrogate loss in (\ref{surrogate_loss}) yields the same change in loss over the parameter displacement $\Delta_k$ as the loss function of the previous tasks.

The path integral over the entire trajectory of parameters during training, however, can yield an inaccurate importance measure when the loss function is not convex. As the loss function of neural networks is generally not convex, we propose to apply exponential decay to $\omega_k$ and $\Delta_k$:
\begin{equation}
\label{path_int}
\omega_k = \lambda \cdot \omega_k - g_k(\theta(t))\Delta_k(t)
\end{equation}
\begin{equation}
\Delta_k = \lambda \cdot \Delta_k + \Delta_k(t)
\end{equation}
where $0 < \lambda < 1$ is a decay factor.

\section{Experiments}
\label{sec:experiments}
\paragraph{Data}
\newcolumntype{C}[1]{>{\centering\arraybackslash}p{#1}}
\begin{table*}[ht!]
\centering
\begin{tabular}{|c||c|c|c|c|}
\hline
Train Data	& \# Dialogs	& Avg. Dialog Len	& Avg. User Len	& Avg. System Len\\ \hline
\texttt{open\_close}	& 10	& 2	& 1.75	& 5.25	\\ \hline
\texttt{HH\_reset\_password}	& 746	& 12.84	& 9.35	& 21.06	\\ \hline
\texttt{HC\_reset\_password}	& 520	& 1.93	& 8.14	& 12.57	\\ \hline\hline
Test Data	& \# Dialogs	& Avg. Dialog Len	& Avg. User Len	& Avg. System Len\\ \hline
\texttt{HC\_reset\_password}	& 184	& 1.99	& 8.33	& 12.22	\\ \hline
\texttt{HC\_reset\_password+}	& 184	& 3.99	& 5.23	& 8.86	\\ \hline
\end{tabular}
\caption{Data statistics}
\label{tab:data}
\end{table*}

In order to test our method, we used four dialog datasets --- \texttt{open\_close}, \texttt{HH\_reset\_password}, \texttt{HC\_reset\_password} and \texttt{HC\_reset\_password+}. ~\footnote{The datasets are not publicly available, but we are unaware of suitable H-H and H-C paired data in the public domain.} Basic statistics of the datasets are shown in Table~\ref{tab:data}. Example dialogs are provided in the Appendix.
\paragraph{\texttt{open\_close}:}
A synthetic corpus that we created in order to clearly demonstrate the phenomena of catastrophic forgetting and the impact of our method. In \texttt{open\_close}, every conversation has only opening/closing utterances without any task-related exchanges. 
\paragraph{\texttt{HH\_reset\_password}:}
A real H-H conversation dataset that we obtained from our company's text-based customer support chat system. All conversation logs are anonymized and filtered by a particular problem which is ``reset password" for this study. Data from this system is interesting in that H-H conversations typically cover not only task-related topics but also various out-of-domain topics including general conversations. If we can transfer such conversational skills to an H-C conversation model, the H-C model will be able to handle many situations even without seeing relevant examples in its training data. But using H-H dialogs to bootstrap a task-oriented dialog system has been shown to be difficult
even with serious annotation effort~\cite{bangalore2008learning}.
\paragraph{\texttt{HC\_reset\_password}:}
A real corpus of H-C conversations that we obtained from our company's text-based customer support dialog system. This data is distinctive since it is used by real users and the system was developed by customer support professionals at our company with sophisticated rules. As the data was obtained from a dialog system, however, it includes mistakes that system made. Thus, we modified the dialog data to correct system's mistakes by replacing it with the most appropriate system action given the context and then discarding the rest of the dialog, since we do not know how the user would have replied to this new system action. The resulting dataset is a mixture of complete and partial conversations, containing only correct system actions. 
\paragraph{\texttt{HC\_reset\_password+}:}
The \texttt{HC\_reset\_password} dataset barely has general conversations since the dialog system for a particular problem only starts after customers are routed to the system from somewhere else according to the brief problem description. But for a standalone system, it is natural to have opening and closing utterances. Thus, we extended the \texttt{HC\_reset\_password} dataset with opening and closing utterances which were randomly sampled from the \texttt{open\_close} dataset. This allows us to test if our method can keep the conversation skills that it learned from either \texttt{open\_close} or \texttt{HH\_reset\_password} after we further train the model on \texttt{HC\_reset\_password}. 


\paragraph{Training Details}
To implement the state tracker, we use three LSTM-RNNs with -- 25 hidden units for each direction of the RNNs for word encoding, 128 hidden units for each direction of the RNNs for utterance encoding and 256 hidden units for the RNN for state encoding. We shared the RNNs for word and utterance encoding for both user utterances and (candidate) system actions. We initialized all LSTM-RNNs using orthogonal weights~\cite{saxe2013exact}. We initialized the character embeddings with 8-dimensional vectors randomly drawn from the uniform distribution $U(−0.01, 0.01)$ while the word embedding weight matrix is initialized with the GloVe embeddings with 100 dimension~\cite{pennington2014glove}. 
We use the Adam optimizer~\cite{kingma2014adam}, with gradients computed on mini-batches of size 1 and clipped with norm value 5. The learning rate was set to $1 \times 10^{-3}$ throughout the training and all the other hyper parameters were left as suggested in \newcite{kingma2014adam}. 
To create train sets for the Plackett-Luce model, we performed negative sampling for each dataset to generate 9 distractors for each truth action.

\paragraph{Continual Learning Details}
As a simple transfer mechanism, we initialized all weight parameters with prior weight parameters when there is a prior model. Given that our focus is few-shot learning, we don't assume the existence of development data with which we can decide when to stop training. Instead, training terminates after 100 epochs which is long enough to reconstruct all training examples. The $\omega_k$ and $\Delta_k$ are updated continuously during training, whereas the importance measure $\Omega_k$ and the prior weight $\tilde{\theta}$ are only updated at the end of each task. We set the trade-off parameter $c$ to $0.01$.
If the path integral (Eq.~\ref{path_int}) is exact, $c = 1$ would mean an equal weighting among tasks. However, the evaluation of the integral typically involves various noises, leading to an overestimate of $\omega_k$. To compensate the overestimation, $c$ generally has to be chosen smaller than one.

\paragraph{Results on Synthetic Dialog to H-C Dialog}
\begin{table*}[ht!]
\centering
\begin{tabular}{|c||C{1.2cm}|C{1.2cm}|C{1.2cm}||C{1.2cm}|C{1.2cm}|C{1.2cm}|}
\hline
Test Data	& \multicolumn{3}{|c||}{\texttt{HC\_reset\_password}}	& \multicolumn{3}{|c|}{\texttt{HC\_reset\_password+}} \\ \hline
\diagbox{Train Size}{Model}	& NT	& WT	& AEWC	& NT	& WT	& AEWC \\ \hline
1	& 47.32	& 49.85	& 50.73	& 18.02	& 24.01	& \bf{27.17} \\ \hline
2	& 54.45	& 55.52	& 55.38	& 22.28	& 22.33	& \bf{35.79} \\ \hline
3	& 58.91	& 60.41	& 58.86	& 25.70	& 23.48	& \bf{47.65} \\ \hline
4	& 61.88	& 61.38	& 60.13	& 27.03	& 23.99	& \bf{49.71} \\ \hline
5	& 63.03	& 62.51	& 60.32	& 27.93	& 22.84	& \bf{51.14} \\ \hline
\end{tabular}
\caption{Experimental results on conversational skill transfer from synthetic to human-computer dialogs.}
\label{tab:res_syn}
\end{table*}

In order to clearly demonstrate the catastrophic forgetting problem, we compare three models trained by different training schemes: 1) {\em No Transfer} (NT) -- we train a model on \texttt{HC\_reset\_password} from scratch, 2) {\em Weight Transfer} (WT) -- we train a model on \texttt{open\_close}, and continued to train the model on \texttt{HC\_reset\_password}, 3) AEWC -- the same as 2) except for AEWC being applied. We compared accuracy on both the evaluation data from \texttt{HC\_reset\_password} and \texttt{HC\_reset\_password+}. 
The result is shown in Table~\ref{tab:res_syn}. All the models perform well on \texttt{HC\_reset\_password} due to the similarity between the training and evaluation data. But the performances of NT and WT on \texttt{HC\_reset\_password+} significantly drop down. This surprisingly poor result confirms ~\newcite{shalyminov2017challenging} which found that neural conversation models can be badly affected by systemic noise. In this case, we systemically introduced unseen turns into dialogs. On the contrary, AWEC shows a higher performance than the others by trying to find optimal parameters not only for the previous task but also for the new task. One of key observations is that the performance of WT on \texttt{HC\_reset\_password+} starts strong but keeps decreasing as more training examples are given. This indicates that weight transfer alone cannot help carry prior knowledge to a new task, rather it might lead to poor local optima if the prior knowlege is not general enough.



\paragraph{Results on H-H Dialog to H-C Dialog} 
\begin{table*}[ht!]
\centering
\begin{tabular}{|c||C{1.2cm}|C{1.2cm}|C{1.2cm}||C{1.2cm}|C{1.2cm}|C{1.2cm}|}
\hline
Test Data	& \multicolumn{3}{|c||}{\texttt{HC\_reset\_password}}	& \multicolumn{3}{|c|}{\texttt{HC\_reset\_password+}} \\ \hline
\diagbox{Train Size}{Model}	& NT	& WT	& AEWC	& NT	& WT	& AEWC \\ \hline
1	& 45.67	& 47.87	& 48.07	& 16.70	& 44.63	& \bf{50.10} \\ \hline
2	& 54.09	& 56.15	& 56.05	& 22.79	& 51.93	& \bf{63.31} \\ \hline
3	& 58.87	& 60.88	& 60.60	& 25.62	& 54.89	& \bf{68.88} \\ \hline
4	& 62.86	& 64.08	& 63.18	& 27.20	& 57.68	& \bf{71.53} \\ \hline
5	& 63.84	& 65.08	& 64.07	& 27.39	& 57.81	& \bf{72.99} \\ \hline
\end{tabular}
\caption{Experimental results on conversational skill transfer from human-human to human-computer dialogs.}
\label{tab:res_hh_hc}
\end{table*}

Now we turn to a more challenging problem --- we bootstrap a conversation model by first learning from H-H dialogs without any manual labeling. We again compare three models trained by different training schemes: 1) {\em No Transfer} -- we train a model on \texttt{HC\_reset\_password} from scratch, 2) {\em Weight Transfer} --  we train a model on \texttt{HH\_reset\_password}, and continued to train the model on \texttt{HC\_reset\_password} 3) AEWC -- the same as 2) except for AEWC being applied.
The characteristics of H-H dialogs are vastly different from H-C dialogs. For example, the average utterance length and conversation length are way longer than H-C dialogs, introducing complex long dependencies. Also, H-H dialogs cover much broader topics that do not exactly match the conversation goal introducing many new vocabularies that do not occur in the corresponding H-C dialogs. 
In order to make the knowledge transfer process more robust to such non-trivial differences, the dropout regularization~\cite{srivastava2014dropout} was applied to both utterance and state embeddings with the ratio of 0.4. We also limited the maximum utterance length to 20 words. 
The result is shown in Table~\ref{tab:res_hh_hc}, which generally agrees on the previous experimental results obtained with synthetic data. All three models work well on \texttt{HC\_reset\_password}. NT performs poorly on \texttt{HC\_reset\_password+} as expected whereas WT interestingly shows a much higher performance than NT. This can be attributed to the fact that \texttt{HH\_reset\_password} contains a broader array of conversation skills including some knowledge on the target task compared to \texttt{open\_close}, which leads to a more general and transferable model that would not necessarily be forgotten when faced with a new task. But the increasing performance gap between WT and AEWC on \texttt{HC\_reset\_password+} clearly shows that WT gradually forgets the prior knowledge as it trains on more examples from a new task. Another noteworthy observation is that, when just one training example is given, the performances of WT and AEWC are much higher than that of NT. This result demonstrates that our method can capture meaningful knowledge from prior tasks in such a way that it partly works on a related new task.

\section{Conclusion}
\label{sec:conclusion}
We have presented a novel approach to the data-intensiveness problem due to the lack of compositionality of end-to-end neural conversation models. We tackled the problem with a continual learning technique for neural networks. Specifically, we proposed a universal conversational model that can be used across different tasks and a novel {\em Adaptive Elastic Weight Consolidation} method which together allows us to continuously train on a sequence of tasks without forgetting earlier tasks. We tested our method with two experimental configurations -- conversation skill transfer from synthetic dialogs/H-H dialogs to H-C dialogs in a customer support domain. Future work includes in-depth analysis on how our algorithm distributes different pieces of knowledge across the network. Also, it would be interesting to apply the continual learning method for composing multiple tasks to handle a more complex task.

\bibliography{acl2017}
\bibliographystyle{acl_natbib}

\appendix
\noindent{\bf Appendices}
\section{\texttt{open\_close} example dialog}
Hello.\\
Hello. What can I help you?\\
Okay. Thank you.\\
Sure thing! Have a great day.

\section{\texttt{HC\_reset\_password} example dialog}
I forgot my password and now I have locked my account for 30 days\\
Okay, you don't need to remember your password, we can reset it. Would you like to try that?\\
Yes please \\
SOLUTION: To reset your password, go to xx\_url\_xx. Was that helpful? \\
Okay that's also another problem I have. I added extra security \& the number I have on  fine is no longer active. I chnsged my number and because I can't access it I can't long into my account \& it won't let me reset my password \\
Let's connect you to a person who can help you.

\section{\texttt{HC\_reset\_password+} example dialog}
Good morning.\\
Hi. What can I help you?\\
forgot password\\
Okay, you don't need to remember your password, we can reset it. Would you like to try that?\\
already tried that\\
Let's connect you to a person who can help you.\\
Okay thank you.\\
Sure thing! Have a great day.

\section{\texttt{HH\_reset\_password} example dialog}
hi, thanks for visiting answer desk! i'm xx\_firstname\_xx q.\\
hello i am having trouble accessing my laptop\\
hi, how may i help you today?\\
i forgot the password i changed the pw on my account, but the computer is still not able to be accessed\\
oh that's bad , that might be very important to you but no worries i will help you out with your issue. to start with may i have your complete name please?\\
my name is xx\_firstname\_xx xx\_firstname\_xx, i am contacting you on behalf of xx\_firstname\_xx mine\\
that's okay xx\_firstname\_xx. may i also know your email address and phone number please?\\
my email, xx\_email\_xx, xx\_phonenumber\_xx\\
thank you for the information. xx\_firstname\_xx what is your current operating system?\\
windows 10\\
may i know what is the error message you received upon trying to unlock your computer?\\
the password is not working i forgot the pw, i tried to reset the pw from the account\\
is it a local account or microsoft account?\\
i am not sure i thought it was a microsoft account but the pw didnt change\\
can you send me the email so that i can check if it is microsoft account?\\
xx\_email\_xx though i think i may have created one by accident setting up the computer it might be xx\_email\_xx or some variation of that i chnaged the pw on the xx\_email\_xx but it had no effect\\
since we're unable to know what exactly happening to your computer. i will provide you our technical phone support so that you will be well instructed on what you are going to do to get your computer work again. would that be okay with you?\\
fine\\
one moment please.

\end{document}